\documentclass[final]{cvpr}
\usepackage[utf8]{inputenc}

\usepackage{times}
\usepackage{epsfig}
\usepackage{graphicx}
\usepackage{subcaption}
\graphicspath{{./images/}}
\usepackage{amsmath}
\usepackage{amssymb}
\usepackage[normalem]{ulem}

\usepackage{enumitem}
\usepackage{url}
\usepackage{xcolor}

\definecolor{orange}{rgb}{1.0, 0.49, 0.0}

\usepackage{xcolor}
\usepackage{amsmath}

\definecolor{negativered}{RGB}{128, 0, 0}
\definecolor{positivegreen}{RGB}{0, 128, 0}

\definecolor{blue1}{RGB}{53, 113, 152}
\definecolor{blue2}{RGB}{65, 141, 191}
\definecolor{blue3}{RGB}{102, 164, 205}
\definecolor{blue4}{RGB}{139, 187, 218}

\newif\ifdraft
\draftfalse
\ifdraft
\usepackage{cancel}
\fi

\usepackage[pagebackref=true,breaklinks=true,colorlinks,bookmarks=false]{hyperref}

\def\cvprPaperID{18} 
\def\confYear{CVPR 2021}
\usepackage[square,numbers]{natbib}
\title{
Plants Don't Walk on the Street:\\ Common-Sense Reasoning for Reliable Semantic Segmentation
}

\author{
    \centerline{{Linara Adilova$^{1,\ast}$, Elena Schulz$^{1,\ast}$, Maram Akila$^{1}$, Sebastian Houben$^{1}$,}}\\ 
    \centerline{{Jan David Schneider$^{2}$, Fabian Hüger$^{2}$, Tim Wirtz$^{1}$}} \\
    \centerline{$^1$ Fraunhofer IAIS, $^2$ Volkswagen AG} \\
    \centerline{\small{$\ast$ equal contribution}} \\
    \centerline{\small{\texttt{\{first.last\}@iais.fraunhofer.de}}} \\
    \centerline{\small{\texttt{fabian.hueger@volkswagen.de}}} \\
    \centerline{\small{\texttt{jan.david.schneider@volkswagen.de}}} \\
}
\begin{document}

\maketitle

\begin{abstract}

Data-driven sensor interpretation in autonomous driving can lead to highly implausible predictions as can most of the time be verified with common-sense knowledge. 
However, learning common knowledge only from data is hard and approaches for knowledge integration are an active research area.
We propose to use a partly human-designed, partly learned set of rules to describe relations between objects of a traffic scene on a high level of abstraction.
In doing so, we improve and robustify existing deep neural networks consuming low-level sensor information.
We present an initial study adapting the well-established Probabilistic Soft Logic (PSL) framework to validate and improve on the problem of semantic segmentation. 
We describe in detail how we integrate common knowledge into the segmentation pipeline using PSL and verify our approach in a set of experiments demonstrating the increase in robustness against several severe image distortions applied to the A2D2\footnote{\url{https://www.a2d2.audi/a2d2/en.html}} autonomous driving data set.

\end{abstract}


\section{Introduction}
\begin{figure}[t]
    \centering
    \includegraphics[width=\columnwidth]{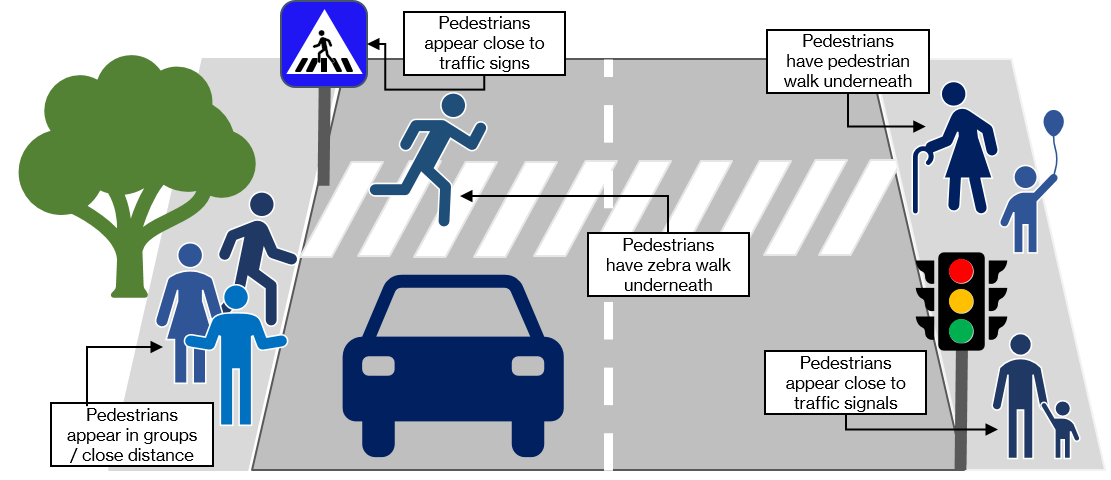}
    \caption{Most humans approach the task of understanding and reasoning about traffic scenes using relations between objects. Given that all of them have been identified, common knowledge --- describing both behaviour and features, \eg relative position --- can straightforwardly be delineated in terms of simple rules. Some of the rules that we utilize in this paper are exemplified in the above figure.}
    \label{fig:psl-motivation}
\end{figure}

Deep neural networks are an integral part of autonomous driving as they achieve unmatched performance in a number of tasks, in particular image-based environment perception.
Still, they are prone to random or targeted disturbances, which renders robustness a crucial requirement for safety-critical applications. 
These disturbances might be of a natural source -- like overexposure, fog, snow -- or of an artificial and purposefully malicious source -- like adversarial attacks.
At the same time, wrong predictions in such tasks are most of the time blatantly obvious for human viewers that use multiple intuitive rules based on their knowledge and experience to reveal what is happening in a given scene.
Neural approaches, in their current form, aim to implicitly learn these rules from data, which makes it necessary to acquire and annotate large amounts of inputs and furthermore introduces the danger of adopting spurious correlations.
The integration of scene knowledge into the training or inference of neural networks seems a viable approach, however, this knowledge ranges from considerations of scene geometry, common knowledge, and rules of human behavior, rendering it a difficult problem \citep{vonrueden2020informed}. 
There are not only a number of competing approaches to perform the integration of knowledge itself but also multiple ways of representing the knowledge.
\par
In this paper, we study a way to represent knowledge in form of relations delineated as logic rules between identified objects in a traffic scene (see \figref{fig:psl-motivation}) and examine how these relations can be used to robustify data-driven environment perception.
To this end, we adapt the Probabilistic Soft Logic (PSL) framework~\citep{kimmig2012short} to define and learn relations between objects in the surrounding scene. 
We demonstrate the results of initial experiments aimed at quantifying its effect on the robustness of a recent small-scale semantic segmentation neural network.

\section{Related Work}
\label{sec:related_work}
The importance of high-level knowledge injection for image-related tasks is widely accepted~\citep{aditya2019integrating}. 
Utilizing background and common-sense knowledge about objects or regions, which goes beyond data annotations, can benefit many computer vision tasks, such as object recognition or semantic segmentation. 
Mostly due to its use in robotics and autonomous driving vehicles, semantic segmentation, as one form of low-level scene representation, is currently one of the actively developing fields in image recognition.
\par
In their survey~\citet{siam2017deep} show the development of semantic segmentation from feature-based approaches to the end-to-end deep learning networks.
Among other techniques, they point out and discuss an increase in accuracy achieved by including structural knowledge, which is abundant in the particular task of autonomous driving. 
To facilitate knowledge inclusion, according to~\citet{siam2017deep}, many researchers apply conditional random fields (CRF) additionally to deep convolutional networks for classification to improve the accuracy of the predictions.
There are also approaches that are based only on an ontology or logical reasoning for semantic segmentation. 
For example, \citet{zand2016ontology} propose to deploy Dirichlet mixture models based on super-pixel segmentation using knowledge from a fixed, user-defined ontology after which prediction is performed using conditional random fields. 
Analogously, \citet{leon2020big} use fuzzy logic to perform segmentation and \citet{dasiopoulou2005knowledge} use a high-level semantic ontology for object detection in videos.
\par
In our work, we propose to employ the PSL framework as an expressive and interpretable way to reason about the environment of the autonomous vehicle. 
To the best of our knowledge it is the first attempt to combine deep-learning-based semantic segmentation with probabilistic logical reasoning about the environment.
Nevertheless, the idea of employing knowledge via logic rules in autonomous driving tasks is not new. 
In fact, \citet{kanaujia2015markov, kardacs2013learning, kembhavi2010did} use Markov logic networks for recognizing complex events in video data.

Related work in the same area of application as ours is by \citet{souza2011probabilistic}, in which the authors use logic reasoning to support a lane detection system with high-level knowledge. 
They employ the output of their segmentation method for further reasoning and controlling the movement of the vehicle (or alerting the driver).
Similar to the PSL framework used in our work, these authors adopt Markov logic networks to employ high-level knowledge in the form of logical hard and soft rules.
However, the object-based rules are only used to classify events after recognizing objects (\eg person, car) within image frames.
In contrast, we employ soft object-based rules to robustify the prediction results of a semantic segmentation neural network for the objects in the image frames itself.
Thus, our goal is to construct rules that analogously would support semantic segmentation in cases when the network itself is compromised (\eg by an image distortion) in order to prevent completely missing critical objects such as pedestrians, traffic signs or traffic signals. 


\section{Approach}
The goal of our research is to find ways in which high-level domain knowledge can improve and robustify the predictions of deep learning models in the context of autonomous driving. 
While multiple lines of research discussed in \secref{sec:related_work} make use of domain knowledge exclusively for building a model, it is our aim to combine the data-driven approach and high-level knowledge during inference.
\par
For various traffic scenes, humans can typically use common and empirical knowledge about behaviour and features of the objects to infer their type and in-scene relations, even if generalizing to unseen scenarios.
Considering different features and interrelations of objects, this human common knowledge can partially be encoded as a set of simple rules, some of which are showcased in \figref{fig:psl-motivation}. 
The way a human would reason about such a traffic scene cannot be based on hard Boolean logic, though, in which rules hold either always or never.
For example, an object having a zebra walk underneath and standing close to a traffic sign has a high probability to be a human, as depicted in \figref{fig:psl-motivation}.
But there is a chance that this could also be something else, \eg a dog or some obstacle in a construction site.
Therefore, to integrate such high-level human knowledge in the same ``uncertain'' manner into a neural network model, we need to define logical rules with relaxed truth values.
Structuring the rules upon the identification and classification of objects aligns with the task of semantic segmentation that lends itself well to finding image regions with a common semantic, which we treat as object proposals.
We selected PSL as a framework for knowledge representation for two reasons: 
First, it allows us to flexibly use soft truth values that are suitable to define both strict and loose rules, and, second, it can be fine-tuned on data, hence, distilling high-level knowledge directly via the given annotations.
\par
We first introduce PSL in \secref{sec:PSL} and briefly describe the input requirements for a PSL program.
Based on this, \secref{sec:DerivingRules} elaborates on how we define object entities and rules in our domain of interest followed by a description of the approach to combine the PSL rules with semantic segmentation.

\subsection{Probabilistic Soft Logic}
\label{sec:PSL}

Probabilistic Soft Logic (PSL)~\citep{Bach2017HLMRFsPSL,kimmig2012short} is a programming framework for probabilistic reasoning in domains with relational structure. 
Typical applications of PSL are collective classification (of interlinked objects in structured data), entity resolution (finding a mapping from multiple interrelated references to the true set of underlying entities) and link prediction (predicting whether a relationship exists between two entities). 
PSL makes use of Hinge-Loss Markov Random Fields (HL-MRF), a graphical model, and applies these to structured machine learning problems. 
Graphical models define a type of loss function that represents a probability score for each possible value configuration respecting the graph-encoded dependencies among the variables. 
For HL-MRFs~\citep{Bach2013HingeLoss,Bach2017HLMRFsPSL} the continuous loss functions are defined using the hinge loss, which allows for efficient convex optimization techniques during inference.
\par
The HL-MRFs are defined using declarative rules, whose importance can be indicated by optional non-negative \textit{weights}. 
These rules are formulated using first order logic.
In contrast to the binary truth values \emph{true} and \emph{false} of Boolean algebra, PSL uses continuous soft truth values in the interval $[0,1]$.
\begin{figure*}[ht]
    \centering
    \includegraphics[width=\textwidth]{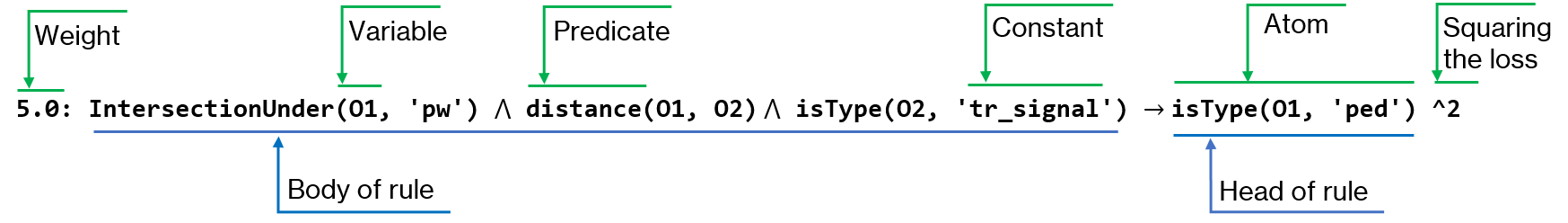}
    \caption{PSL rules have a conjuncted body and a single literal head. \emph{Atoms} (\eg \texttt{IntersectionUnder(O1, 'r')} are the basic elements that are used for reasoning.
    An atom is a combination of \emph{predicates} (here: \texttt{IntersectionUnder$\backslash$2}) and a sequence of variables and constants equal to the predicate's arity.
    A \emph{constant} indicates a concrete element (\eg ``\texttt{ped}'' is the concrete object class pedestrian) and a \emph{variable} denotes a placeholder later to be substituted by concrete elements.
    Here, the predicate \texttt{IntersectionUnder} has an arity of two, thus, the atom consists of the variables \texttt{O1} indicating an object plus the constant ``\texttt{r}'' indicating the underground type is the road.
    The identifier \texttt{O1} will be substituted for all available objects during inference while the constant ``\texttt{r}'' will stay the same.
    The \texttt{O1} identifier is used in three atoms indicating that during inference, each object-1-variable will be substituted by the same constant (same holds for object-2-variable \texttt{O2}, respectively).
    The exponent \texttt{\^}\texttt{2} at the end of the rule indicates that the quadratic loss is chosen. This example rule can be interpreted as \emph{``if an object 1 is placed on a sidewalk and in close distance to another object 2, which is a traffic signal, there is a probability that this object 1 is a pedestrian''}.}
    \label{fig:psl-rule-example}
\end{figure*}
For a detailed explanation of the elements of a PSL rule, see the example depicted in \figref{fig:psl-rule-example}. 
\par
Conceptually, we require that the ``satisfaction'' of rules can be measured in a soft-Boolean way if all ``variables'' have a defined value. In this case, we can aggregate the set of all rules to a global score using individual weights for each rule. 
Given a set of ``ground truth'' observations, the weights can be fine-tuned such that the importance of each rule agrees with the ground truth observations. 
The resulting fixed set of weights then allows us to perform inference over incomplete observations by choosing the most probable value for unknown variables.

In detail, this means each atom is mapped to a soft truth value -- before inference for all known variables, and during inference for all unknown variables.
To compute the continuous truth value of the whole rule, values on atom-level are combined using the \emph{Lukasiewicz t-norm}~\citep{klir1995fuzzy} as a relaxation for the logical conjunction, disjunction, and negation.
Each rule's \emph{satisfaction} is calculated indicating to what degree that rule is fulfilled given the current configuration of values.
A rule is satisfied if the head is assigned at least the same truth value as the body, and a fully satisfied rule has a value of $1$.

All possible variable configurations for the rules are combined to a probability distribution, in which those that satisfy more ground rules are more probable.
%
If a weight is assigned to a rule, this indicates the importance of this rule to be satisfied in relation to the other rules and their corresponding weights. 
Rules that are unweighted induce hard constraints which require the rule to be met at all times.
During inference, the most probable value configuration for unknown predicates is found given evidence (via known predicates).
Given ground truth, the weights of rules can also be learned with maximum-likelihood estimation. 
Thus, the importance of each rule in the rule set can be learned from the data itself.
\par
Apart from these rules, a PSL program consists of a database that defines the elements of the domain of interest as well as their relations in the form of predicates. 
The database is built up from known predicates (\ie atoms that are completely observed) and unknown predicates (\ie atoms that are unobserved and must be inferred) that are part of one of the three following sets:
(i) Necessary for all known predicates is the set of \textit{observations}. These are variable combinations with an associated soft truth value in the form of actual extracted and known data for each.
(ii) Necessary for all unknown predicates is the set of \textit{targets} that is a list of all atoms (combinations of variables) that should be inferred, therefore without any associated truth values.
(iii) For weight learning a set of \textit{truths} with actual known data (likewise observations) to be used as ground truth for all unknown predicates is necessary.
Inference in PSL is done based on the first two sets, while evaluation of the inferred predicates can be done with the test set of the \textit{truths}.

There are two types of rules that are supported in PSL: logical and arithmetic rules. 
We use logical rules to encode dependencies between atoms as for example depicted in the rule in \figref{fig:psl-rule-example}.
Additionally, we need to imply an arithmetic constraint to restrict the substitution for variables further.
In our setting, we want to reason about the class of an object in the scene that can only be one of three classes, constraiing the corresponding truth values to sum up to $1$.

\subsection{Deriving Rules from Domain Knowledge}
\label{sec:DerivingRules}
\par
The overall goal of employing PSL to camera-based perception in autonomous driving is to identify parts of common knowledge about the traffic scene that can be readily extracted and represent it in a form that facilitates integration into the model performing inference.
Based on the input requirements for a PSL program as defined above, we first need a database, \ie objects that we want to reason about as well as their relationships encoded within chosen predicates. 
Then, we define a set of rules that carries additional relational knowledge about the domain of interest that is not yet covered by the annotated data and thus can be deployed to support and validate the perception task. 
When defining an initial set of rules, one therefore has to carefully choose the objects they link up: it should be simple to both describe their behavior and to extract them from raw data.

Here, we choose pedestrians, traffic signs and traffic signals as our three types of objects of interest, for which we want to infer the class.
To identify these objects in a logical setting, one would need to fall back to low-level sensor interpretation.
Typically, autonomous vehicles have diverse sensors from which objects, their attributes (\eg colors, shapes) and their surroundings (\eg area underneath, background) and their relationships (\eg distance to other objects) can be derived. 
\par
For our case study, however, we solely focus on the aspect of how external knowledge can be meaningfully integrated into an existing prediction workflow. 
We therefore purposefully deploy an oracle to extract the object entities thereby neglecting additional challenges resulting from, \eg,  unreliable sensor readings and object extraction.
Our solution is to employ the connected components on ground truth segmentation masks using additional smoothing to avoid fractured objects due to occlusion.
Additionally, we emulate a depth sensor by propagating the RGB inputs to a publicly available\footnote{\url{https://github.com/nianticlabs/monodepth2}} version of MonoDepth2~\citep{godard2019digging}. 
This ``oracle sensor'' information already allows for creating a meaningful set of objects as basis for the rules.

Subsequently, the rules describing the objects have to be defined.
To this end, we develop an initial set of rules, for which we refine the weights later on.
As we want to reason about the objects of our three selected classes, we define our target predicate \texttt{isType(Obj,Class)}.
This is used to later infer a soft truth value for each object and the possible three classes, resulting in one class being more likely than the others.
We then base the logical reasoning for an object class on the semantic features, on the surroundings of the objects, and on their relations.
Surroundings of the objects are described with the observations (known predicates) \texttt{intersectionUnder(Obj,Class)} and \texttt{intersectionBehind(Obj,Class)}.
For example, to infer the class of an object to be ``traffic sign'' \texttt{intersectionUnder} should have a high likelihood for the classes ``pedestrian walk'' or ``street''.
We integrate color as a semantic feature as it is especially important for traffic signs and traffic signals with the predicate \texttt{hasColor(Obj,Color).}
Rules we defined using the color predicate are for example inspired by the officially specified color combinations of common German traffic signs.
The relations between objects are specified as a distance value based on depth readings with the known predicate \texttt{distance(Obj1,Obj2)}.
The defined rules for our setup can be found in appendix~\ref{full_set_rules}.
\par
As a final step, the knowledge gained from the set of data via probabilistic rules has to be combined with the predictions from the neural network, in this case the semantic segmentation masks.
For our conceptual demonstration we follow a simplified approach integrating scores obtained from the network as prior assumptions within the rule based framework.
\begin{figure}[h]
    \centering
    \includegraphics[width=\columnwidth]{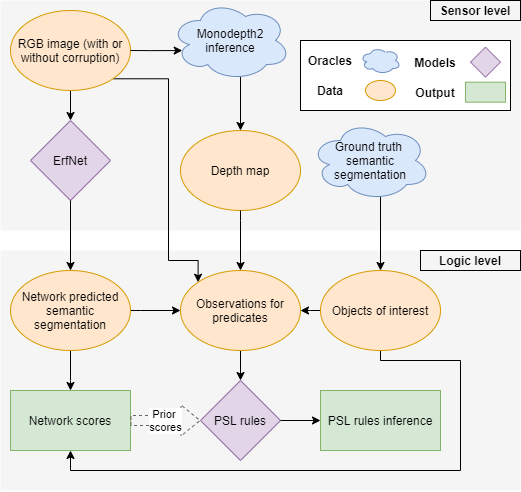}
    \caption{Overall flow of information: Oracles are serving as a source of information for objects and depth maps and also observations for weights learning; predictions produced by the neural network are combined with PSL rules via priors.}
    \label{fig:approach_diagram}
\end{figure}
\par
Overall, the practical approach can be described as follows (see \figref{fig:approach_diagram}):
\begin{enumerate}
    \item Identify objects to reason about and extract them from sensor data (oracle using ground truth semantic segmentation);
    \item Construct the set of rules using expert and common knowledge;
    \item Prepare the required information from lower level sensors for the rules inference (depth, colors);
    \item Learn the weights of the rules using a training set of recorded camera frames;
    \item Perform inference with the neural network and with the rules on the test set of frames for validation, possibly using network outputs as prior.
\end{enumerate}

\section{Evaluation}
The model that we selected for semantic segmentation is ERFNet~\citep{romera2017erfnet} whose lightweight encoder-decoder architecture and limited performance make it straightforward to observe the benefit of knowledge injection via PSL. 
We trained ERFNet on the A2D2 public data set \citep{geyer2020a2d2}, which features a large set of environmental scenes for cars on the road. 
The data set was separated into independent subsets (by the sequences recorded) for training and testing. 
The labels in the data set were simplified to $13$ classes via grouping. 
The network was trained using the Adam optimizer with decaying learning rate for $200$ epochs. 
The network achieved an mIoU of $0.59$ after training--in particular $0.26$ for 'Pedestrian' class and $0.63$ for 'Traffic sign' and 'Traffic signal'.

Rules selected for the evaluation first have to be ranked with weights. 
In our evaluation we obtained the weights using the learning procedure, in which the weights are assigned to the rules based on the training set with ground truth annotations. 
The observations for weight learning are extracted from the training images (predicate \texttt{hasColor}), their depth information (predicate \texttt{distance}) and the ground truth semantic segmentation (predicates \texttt{intersectionBehind} and \texttt{intersectionUnder}) using in each case the object entities from the oracle.
The full list of rules we defined along with the learned weights can be found in appendix~\ref{full_set_rules}. 
This step of weight learning serves to validate our initially proposed rules discarding those which turned out to be inadequate given the data.
In these cases, the weights assigned were in the order of $10^{-5}$. 
One such rule was for example increasing the probability of an object to be a pedestrian if it is close to an object that is a traffic sign. 

An independent test set of images is used both for evaluation of the network prediction and rule inference. 
The connected components selected as objects for soft rule-based reasoning are compared against the network predictions for the same region. 
These regions are produced using the oracle.
We refer to the ratio of correctly classified pixels as the network score for this object. 
Note that for weight learning we used information from the ground truth while for inference network predictions are employed. 

\begin{figure}[th]
    \centering
    \begin{subfigure}{\columnwidth}
        \centering
        \includegraphics[width=0.46\columnwidth]{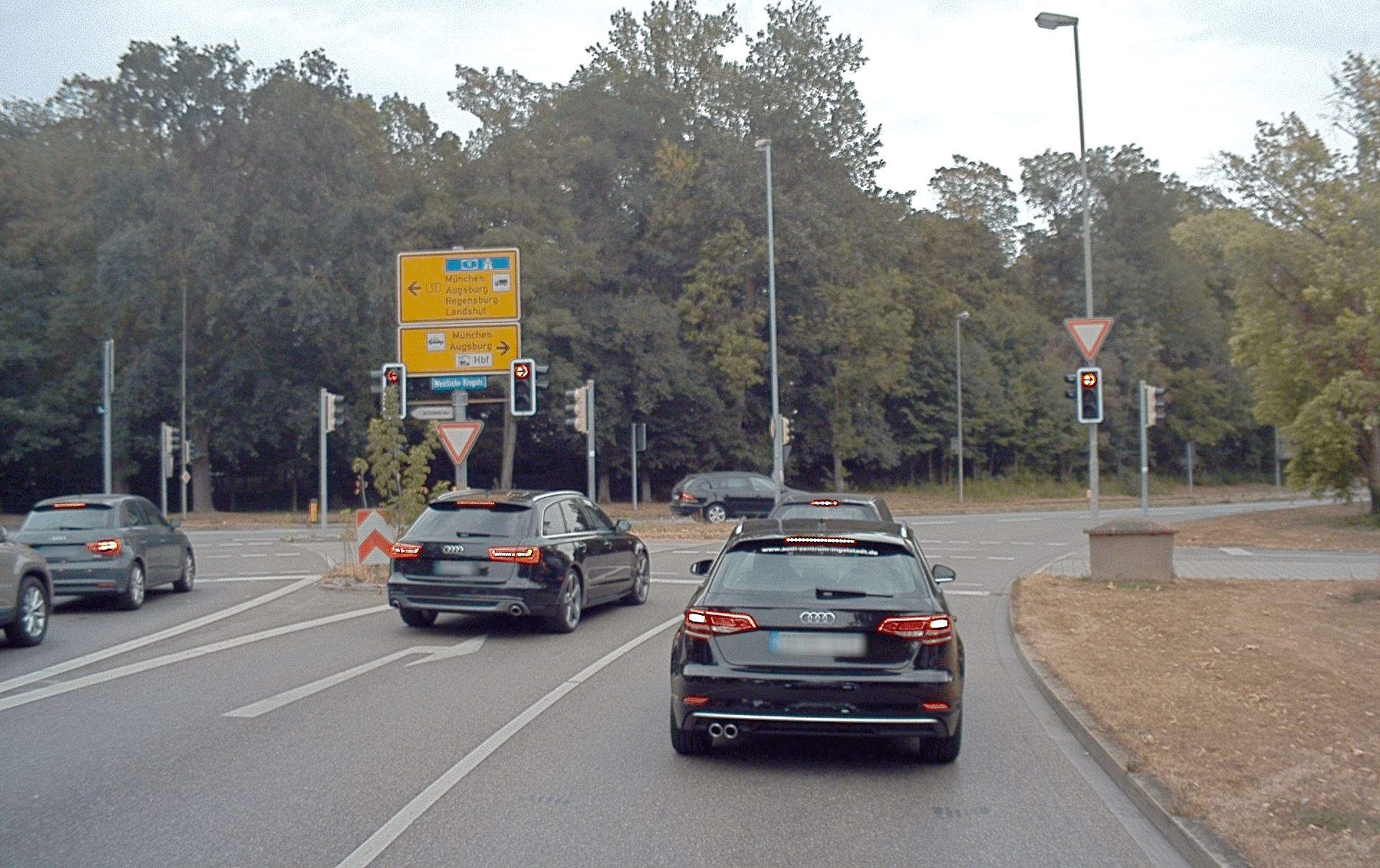}%
        \hfill
        \includegraphics[width=0.46\columnwidth]{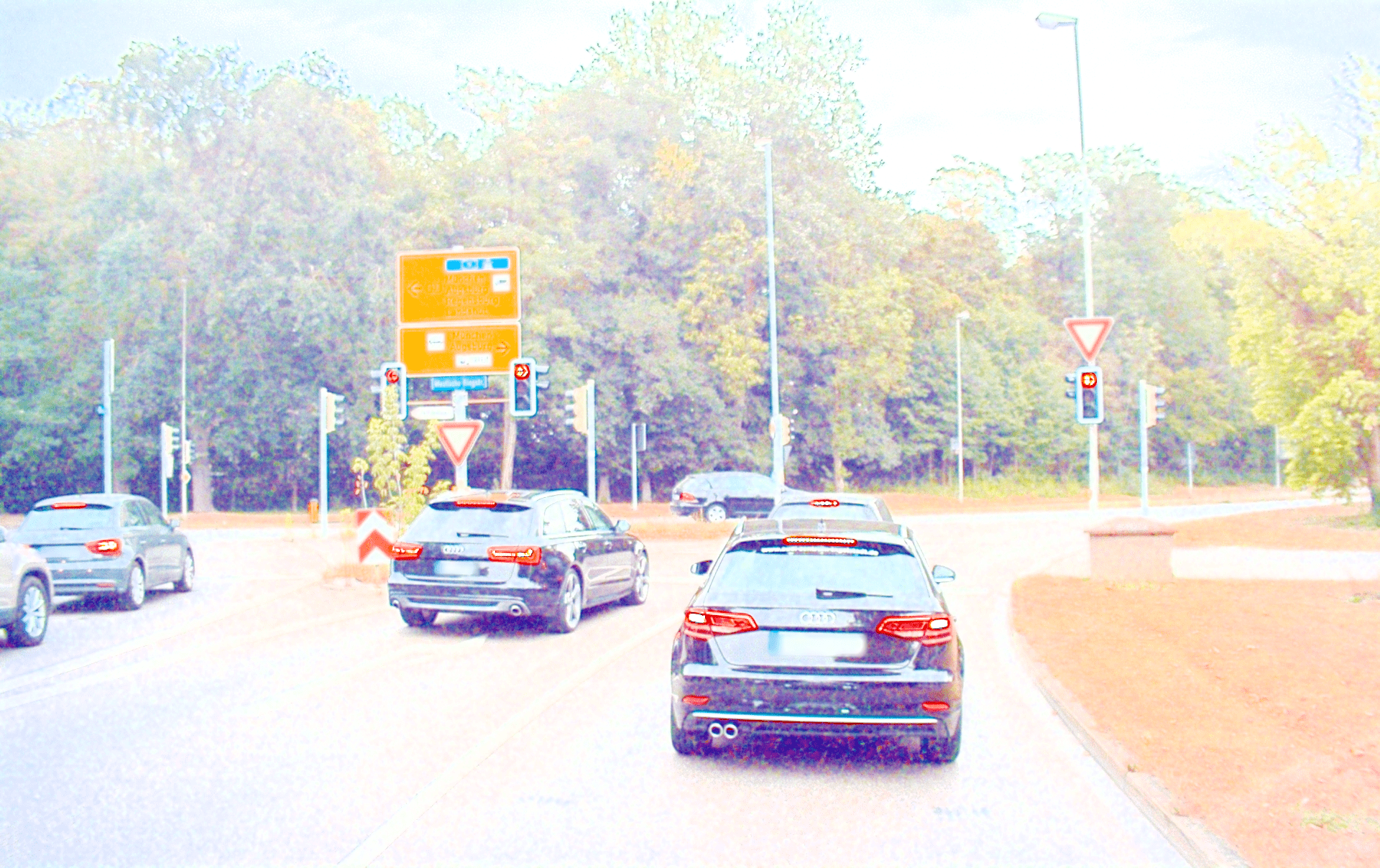}
    \end{subfigure}
    \vskip\baselineskip
    \begin{subfigure}{\columnwidth}
        \centering
        \includegraphics[width=0.46\columnwidth]{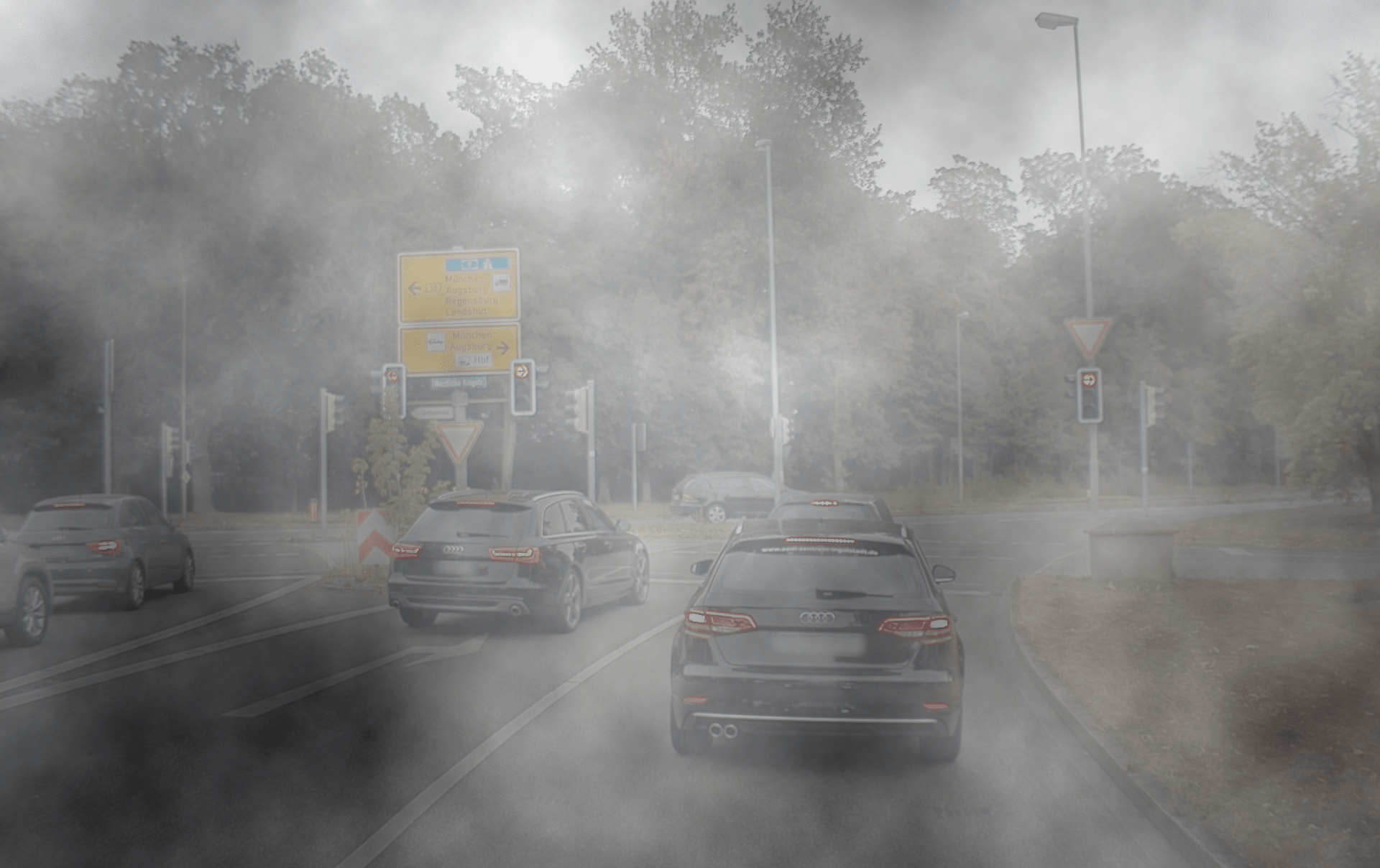}%
        \hfill
        \includegraphics[width=0.46\columnwidth]{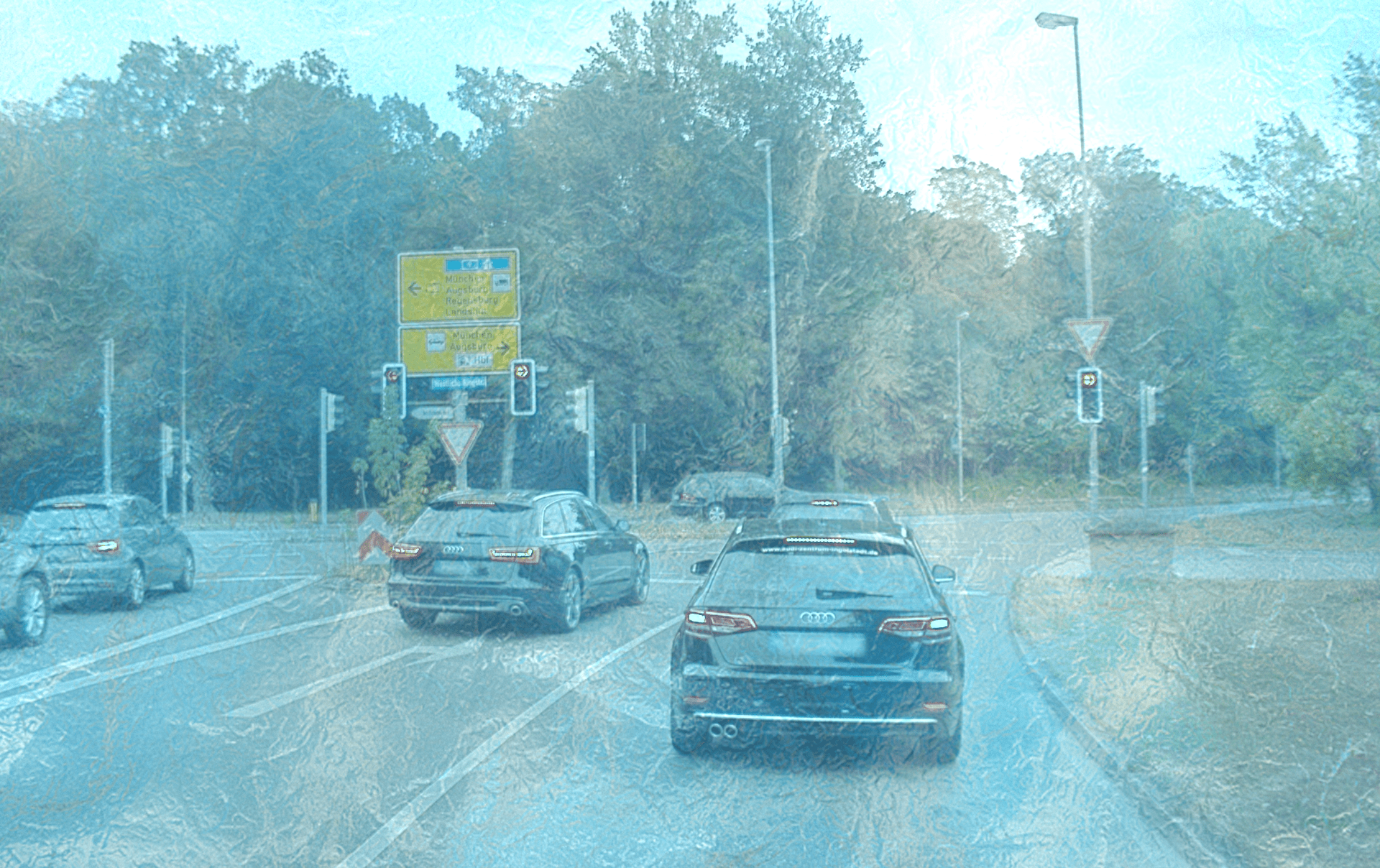}
    \end{subfigure}
    \vskip\baselineskip
    \caption{Example of corruptions: original image (top left), brightness, frost and fog corruption (clockwise).}
    \label{fig:corrupt_image}
\end{figure}

\begin{figure}[h]
    \centering
    \begin{subfigure}[a]{\columnwidth}
        \centering
        \includegraphics[width=0.49\columnwidth]{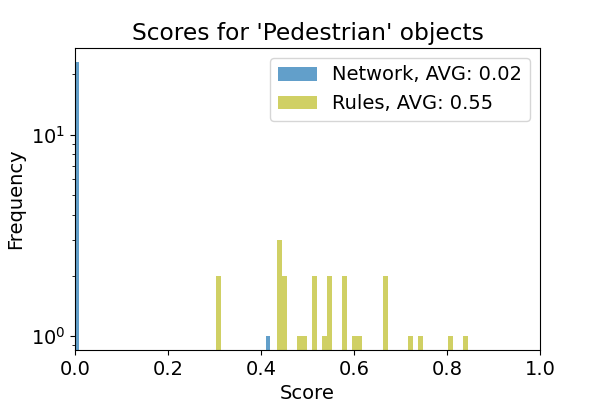}%
        \hfill
        \includegraphics[width=0.49\columnwidth]{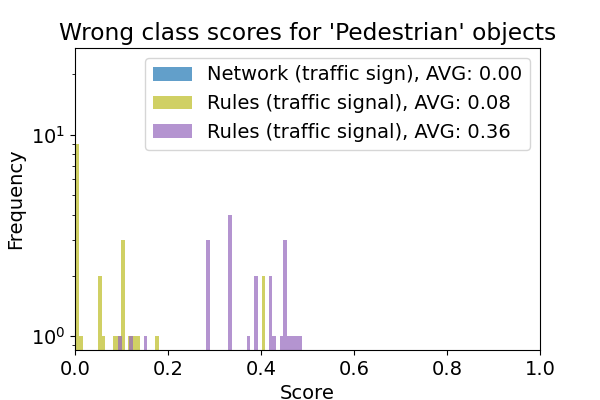}
        \caption{Scores distribution for the objects ``pedestrians'' on the original images.}
    \end{subfigure}
    \vskip\baselineskip
    \begin{subfigure}[b]{\columnwidth}
        \centering
        \includegraphics[width=0.49\columnwidth]{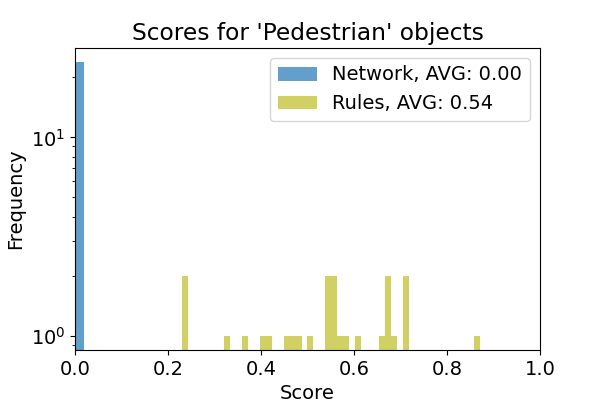}%
        \hfill
        \includegraphics[width=0.49\columnwidth]{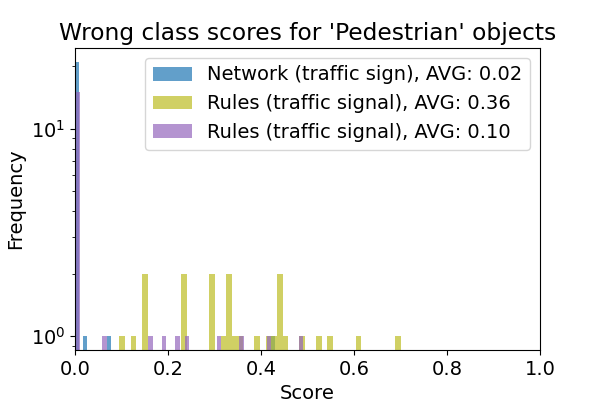}
        \caption{Scores distribution for the objects ``pedestrians'' on the corrupted (brightness, severity 5) images.}
    \end{subfigure}
    \vskip\baselineskip
    \caption{The scores of the network compared to the scores of the rules on the test dataset for pedestrian objects. It should be noted that the distribution of objects is unbalanced in the test set: the number of objects ``pedestrian'' is the smallest, followed by ``traffic signal''. The most frequent class of objects is ``traffic sign''. The network was not trained to distinguish between traffic signs or signals and we thus only report the (joined) result as ``tr\_sign'' for the wrong classes.}
    \label{fig:histograms}
\end{figure}

To evaluate the robustness of the rule predictions we employed image corruptions as described by~\citet{michaelis2019dragon}. 
As the most natural we selected four, namely brightness, fog, frost and snow.
Each of them can be applied with five levels of severity, of which $1$ is the weakest (see \figref{fig:corrupt_image} for examples). 
The evaluation then follows the same route -- the network infers predictions on the corrupted images and the scores for the objects are computed. 
Likewise, rules are inferred on corrupted images and their respective predicted semantic segmentation on them. 
The oracle as further source of information is left unaffected by the applied corruptions.
The results of the predictions are depicted in \figref{fig:histograms}. 
We display the scores for the pedestrian objects, either denoting the correct or incorrect class, in the case of the original images and images corrupted with brightness. 
The histograms for the other two object classes is shown in the appendix~\ref{eval_plots}.
One can observe that the scores produced by the rules are nearly not affected--judging by the distribution of histogram peaks and the average score displayed in the legend--while network scores are very sensitive to the corruption of the data.

\begin{figure}[h]
    \centering
    \includegraphics[width=0.8\columnwidth]{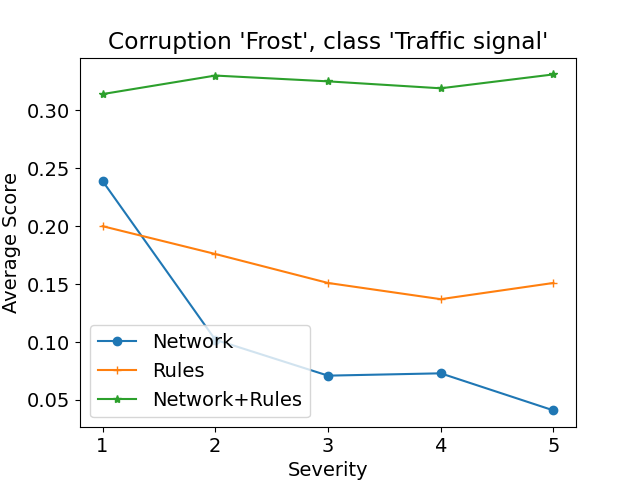}
    \caption{The development of average scores for the class ``traffic signal'' with  increasing severity of the ``frost'' corruption.}
    \label{fig:corruption_vs_score}
\end{figure}

The final evaluation step is to combine network predictions and rules inference. 
We used the scores produced by the network as a prior for the \texttt{isType} predicate in the PSL rules.
\figref{fig:corruption_vs_score} reflects the change in the scores assigned to the objects under increasing severity of corruption. 
The scores for the objects of class ``traffic signal'' are presented here, the other classes and corruptions can be seen in the appendix~\ref{all_corr_plots}.
One can observe that the severity of the corruption affects the scores produced by a network while the inference with the rules deteriorates only slightly. 
Finally, the robustification of the network scores is also stable and shows better results than network scores on their own.

\section{Discussion and Conclusion}

We propose a way to reason about high level scene properties, \ie the type or class of objects, in terms of Probabilistic Soft Logic (PSL).
Taking advantage of human knowledge, we draw information from various domains, ranging from  sensor input such as color to scene properties such as the relations and distances among object.
This approach  enables us to  include additional information from data-driven model predictions, \eg as prior assumption, and improve their prediction reliability via reasoning.
Our initial evaluation in a simplified setup features a scenario from autonomous driving and indicates that the approach is viable.
Even in tests with highly adverse input augmentations it is possible to robustify the predictions of the employed semantic segmentation network.


Within the PSL framework reasoning is based on rules, \ie human understandable concepts regarding the interpretation and meaning of object relations or properties.
Besides the possibility to guide the rule creation by human expert knowledge this offers strong interpretability, the set of rules could, \eg, be audited with respect to consistency.
The relevance of each rule can be adjusted individually, where the necessary weights are either deliberately chosen or learned from a training data set.
The latter case, as demonstrated in our evaluation, can serve to validate the applicability of given rules.
This places PSL in-between human designed and data driven approaches, making it an especially suited candidate to incorporate common knowledge into fundamentally data driven applications.

This work showcases the potential benefits such common knowledge can have for the reliability of a machine learning framework.
At this early stage we, however, neglected the cumbersome step to obtain the entities the PSL framework reasons about.
In practice this information has to be aggregated from various, most likely not fully reliable, sources, for instance stemming from additional sensors (\eg LIDAR) or alternative processing approaches.
While a probabilistic framework seems appropriate to deal with further uncertain input we restrict this demonstration to only a single ML model providing the semantic segmentation mask.
The entities and their further attributes are instead given by an oracle using the ground truth annotation.
This allows us to test the approach without the necessity to instantiate a full framework.
Another important step is the involvement of human experts for rules formulation. The presented rules are formulated with the common knowledge of the driving situations and environment, but setup specific rules, as well as the rules that are tuning a particular aspect of recognition need multiple iterations over the rules construction process.

With respect to the incorporation of machine learning models we currently only use the provided prediction.
Given the probabilistic nature of both these outputs and the PSL framework the inclusion of prediction uncertainty in terms of, \eg, calibrated softmax scores could improve the reliability further.
Adding further models as discussed above would additionally require a larger set of rules.
While the used set is sufficient for demonstration purposes enlarging it to cover more attributes seems to be a straightforward way to extend our work.
Specific follow up questions could address a more systematic approach to rule building, which investigates the interplay between the set coverage, reflected by the amount and variety of rules, and its achieved performance, reflected in learned weights and task performance, more closely.
For the case of autonomous driving the incorporation of temporal consistency, requiring, \eg, that a pedestrian on two consecutive frames should not unexpectedly vanish, could add new aspects to the discussion, possibly including reasoning on the trajectories and intentions of road users.

Especially for domains such as automated driving we believe that the inclusion of human expertise into predictions is a valuable addition to safety and reliability.
Often humans are capable to infer relevant information from context.
For instance a sign indicating the potential presence of children could, by formulation of an appropriate rule, increase the likelihood that a discovered entity is inferred as child.
This inclusion of external knowledge might additionally help counterbalance the rareness of safety relevant situations, which would otherwise make it difficult to learn relevant correlations purely from data.
Deliberately fixing weights or the inclusion of specific rules as above might therefore increase the safety of vulnerable road users (VRUs) without affecting the training and optimization of the underlying machine learning models.

\paragraph{Acknowledgements}
The authors want to thank for support and useful discussions Birgit Kirsch, Julia Rosenzweig, and Dorina Weichert.
\par
The work of L.A. was supported by the Fraunhofer Center for Machine Learning within the Fraunhofer Cluster for Cognitive Internet Technologies. 
The research of S.H. was partly funded by the Federal Ministry of Education and Research of Germany as part of the competence center for machine learning ML2R (01IS18038B).

{\small
\bibliography{bibliography}

\begin{thebibliography}{18}
\providecommand{\natexlab}[1]{#1}
\providecommand{\url}[1]{\texttt{#1}}
\expandafter\ifx\csname urlstyle\endcsname\relax
  \providecommand{\doi}[1]{doi: #1}\else
  \providecommand{\doi}{doi: \begingroup \urlstyle{rm}\Url}\fi

\bibitem[Aditya et~al.(2019)Aditya, Yang, and Baral]{aditya2019integrating}
S.~Aditya, Y.~Yang, and C.~Baral.
\newblock Integrating knowledge and reasoning in image understanding.
\newblock In \emph{Proceedings of the Twenty-Eighth International Joint
  Conference on Artificial Intelligence}, 2019.

\bibitem[Bach et~al.(2013)Bach, Huang, London, and Getoor]{Bach2013HingeLoss}
S.~H. Bach, B.~Huang, B.~London, and L.~Getoor.
\newblock Hinge-loss markov random fields: Convex inference for structured
  prediction.
\newblock \emph{CoRR}, abs/1309.6813, 2013.
\newblock URL \url{http://arxiv.org/abs/1309.6813}.

\bibitem[Bach et~al.(2017)Bach, Broecheler, Huang, and
  Getoor]{Bach2017HLMRFsPSL}
S.~H. Bach, M.~Broecheler, B.~Huang, and L.~Getoor.
\newblock Hinge-loss markov random fields and probabilistic soft logic.
\newblock \emph{J. Mach. Learn. Res.}, 18\penalty0 (1):\penalty0 3846–3912,
  Jan. 2017.
\newblock ISSN 1532-4435.

\bibitem[Dasiopoulou et~al.(2005)Dasiopoulou, Mezaris, Kompatsiaris,
  Papastathis, and Strintzis]{dasiopoulou2005knowledge}
S.~Dasiopoulou, V.~Mezaris, I.~Kompatsiaris, V.-K. Papastathis, and M.~G.
  Strintzis.
\newblock Knowledge-assisted semantic video object detection.
\newblock \emph{IEEE Transactions on Circuits and Systems for Video
  Technology}, 15\penalty0 (10):\penalty0 1210--1224, 2005.

\bibitem[Geyer et~al.(2020)Geyer, Kassahun, Mahmudi, Ricou, Durgesh, Chung,
  Hauswald, Pham, Mühlegg, Dorn, Fernandez, Jänicke, Mirashi, Savani, Sturm,
  Vorobiov, Oelker, Garreis, and Schuberth]{geyer2020a2d2}
J.~Geyer, Y.~Kassahun, M.~Mahmudi, X.~Ricou, R.~Durgesh, A.~S. Chung,
  L.~Hauswald, V.~H. Pham, M.~Mühlegg, S.~Dorn, T.~Fernandez, M.~Jänicke,
  S.~Mirashi, C.~Savani, M.~Sturm, O.~Vorobiov, M.~Oelker, S.~Garreis, and
  P.~Schuberth.
\newblock A2d2: Audi autonomous driving dataset, 2020.

\bibitem[Godard et~al.(2019)Godard, Mac~Aodha, Firman, and
  Brostow]{godard2019digging}
C.~Godard, O.~Mac~Aodha, M.~Firman, and G.~J. Brostow.
\newblock Digging into self-supervised monocular depth estimation.
\newblock In \emph{Proceedings of the IEEE/CVF International Conference on
  Computer Vision}, pages 3828--3838, 2019.

\bibitem[Kanaujia et~al.(2014)Kanaujia, Choe, and Deng]{kanaujia2015markov}
A.~Kanaujia, T.~E. Choe, and H.~Deng.
\newblock Complex events recognition under uncertainty in a sensor network.
\newblock \emph{CoRR}, abs/1411.0085, 2014.
\newblock URL \url{http://arxiv.org/abs/1411.0085}.

\bibitem[Karda{\c{s}} et~al.(2013)Karda{\c{s}}, Ulusoy, and
  {\c{C}}i{\c{c}}ekli]{kardacs2013learning}
K.~Karda{\c{s}}, {\.I}.~Ulusoy, and N.~K. {\c{C}}i{\c{c}}ekli.
\newblock Learning complex event models using markov logic networks.
\newblock In \emph{2013 IEEE International Conference on Multimedia and Expo
  Workshops (ICMEW)}, pages 1--6. IEEE, 2013.

\bibitem[Kembhavi et~al.(2010)Kembhavi, Yeh, and Davis]{kembhavi2010did}
A.~Kembhavi, T.~Yeh, and L.~S. Davis.
\newblock Why did the person cross the road (there)? scene understanding using
  probabilistic logic models and common sense reasoning.
\newblock In \emph{European Conference on Computer Vision}, pages 693--706.
  Springer, 2010.

\bibitem[Kimmig et~al.(2012)Kimmig, Bach, Broecheler, Huang, and
  Getoor]{kimmig2012short}
A.~Kimmig, S.~Bach, M.~Broecheler, B.~Huang, and L.~Getoor.
\newblock A short introduction to probabilistic soft logic.
\newblock In \emph{Proceedings of the NIPS Workshop on Probabilistic
  Programming: Foundations and Applications}, pages 1--4, 2012.

\bibitem[Klir and Yuan(1995)]{klir1995fuzzy}
G.~Klir and B.~Yuan.
\newblock \emph{Fuzzy Sets and Fuzzy Logic: Theory and Applications}.
\newblock Prentice Hall PTR, 1995.
\newblock ISBN 9780131011717.

\bibitem[Leon-Garza et~al.(2020)Leon-Garza, Hagras, Pe{\~n}a-Rios, Conway, and
  Owusu]{leon2020big}
H.~Leon-Garza, H.~Hagras, A.~Pe{\~n}a-Rios, A.~Conway, and G.~Owusu.
\newblock A big bang-big crunch type-2 fuzzy logic system for explainable
  semantic segmentation of trees in satellite images using hsv color space.
\newblock In \emph{2020 IEEE International Conference on Fuzzy Systems
  (FUZZ-IEEE)}, pages 1--7. IEEE, 2020.

\bibitem[Michaelis et~al.(2019)Michaelis, Mitzkus, Geirhos, Rusak, Bringmann,
  Ecker, Bethge, and Brendel]{michaelis2019dragon}
C.~Michaelis, B.~Mitzkus, R.~Geirhos, E.~Rusak, O.~Bringmann, A.~S. Ecker,
  M.~Bethge, and W.~Brendel.
\newblock Benchmarking robustness in object detection: Autonomous driving when
  winter is coming.
\newblock \emph{arXiv preprint arXiv:1907.07484}, 2019.

\bibitem[Romera et~al.(2017)Romera, Alvarez, Bergasa, and
  Arroyo]{romera2017erfnet}
E.~Romera, J.~M. Alvarez, L.~M. Bergasa, and R.~Arroyo.
\newblock Erfnet: Efficient residual factorized convnet for real-time semantic
  segmentation.
\newblock \emph{IEEE Transactions on Intelligent Transportation Systems},
  19\penalty0 (1):\penalty0 263--272, 2017.

\bibitem[Siam et~al.(2017)Siam, Elkerdawy, Jagersand, and
  Yogamani]{siam2017deep}
M.~Siam, S.~Elkerdawy, M.~Jagersand, and S.~Yogamani.
\newblock Deep semantic segmentation for automated driving: Taxonomy, roadmap
  and challenges.
\newblock In \emph{2017 IEEE 20th international conference on intelligent
  transportation systems (ITSC)}, pages 1--8. IEEE, 2017.

\bibitem[Souza and Santos(2011)]{souza2011probabilistic}
C.~R. Souza and P.~E. Santos.
\newblock Probabilistic logic reasoning about traffic scenes.
\newblock In \emph{Conference Towards Autonomous Robotic Systems}, pages
  219--230. Springer, 2011.

\bibitem[von Rueden et~al.(2020)von Rueden, Mayer, Beckh, Georgiev,
  Giesselbach, Heese, Kirsch, Pfrommer, Pick, Ramamurthy, Walczak, Garcke,
  Bauckhage, and Schuecker]{vonrueden2020informed}
L.~von Rueden, S.~Mayer, K.~Beckh, B.~Georgiev, S.~Giesselbach, R.~Heese,
  B.~Kirsch, J.~Pfrommer, A.~Pick, R.~Ramamurthy, M.~Walczak, J.~Garcke,
  C.~Bauckhage, and J.~Schuecker.
\newblock Informed machine learning -- a taxonomy and survey of integrating
  knowledge into learning systems, 2020.

\bibitem[Zand et~al.(2016)Zand, Doraisamy, Halin, and
  Mustaffa]{zand2016ontology}
M.~Zand, S.~Doraisamy, A.~A. Halin, and M.~R. Mustaffa.
\newblock Ontology-based semantic image segmentation using mixture models and
  multiple crfs.
\newblock \emph{IEEE Transactions on Image Processing}, 25\penalty0
  (7):\penalty0 3233--3248, 2016.

\end{thebibliography}
}

\newpage
\onecolumn
\section{Appendix}
\subsection{Full Set of Rules}
\label{full_set_rules}
Predicates:
\begin{itemize}
    \item INTERSECTIONUNDER(object, class, image) - Known predicate (observation). Describes if the area under the object is of a particular class in the image. The classes are 'r' - road, 'pw' - pedestrian walk, 'gr' - ground. During rule weight optimization obtained from ground truth semantic segmentation, during inference obtained from the predictions of a network.
    \item INTERSECTIONBEHIND(object, class, image) - Known predicate (observation). Describes if the area behind the object is of particular class in the image. The class is 'bckgr' - combined buildings and nature. During rule weight optimization obtained from ground truth semantic segmentation, during inference obtained from the predictions of a network.
    \item DISTANCE(object1, object2, image) - Known predicate (observation). Describes if two objects are close to each other in the image. Inferred from the RGB input and corresponding depth map.
    \item HASCOLOR(object, color, image) - Known predicate (observation). Describes if the object has the color in it. The colors are 'white', 'blue', 'yellow', 'red'. Inferred from the RGB input.
    \item ISTYPE(object, class, image) - Unknown predicate (target). Defines the class of the object. Class can be one of 'ped' (pedestrian), 'tr\_sign' (traffic sign), 'tr\_signal' (traffic signal)
\end{itemize}

\begin{small}

\begin{verbatim}
0.768: INTERSECTIONUNDER(O1, 'r', I) >> ISTYPE(O1, 'ped', I) ^2

0.957: INTERSECTIONUNDER(O1, 'pw', I) >> ISTYPE(O1, 'ped', I) ^2

0.789: INTERSECTIONUNDER(O1, 'gr', I) >> ISTYPE(O1, 'ped', I) ^2

3.2E-5: INTERSECTIONBEHIND(O1, 'bckgr', I) >> ISTYPE(O1, 'ped', I) ^2

0.234: ~INTERSECTIONBEHIND(O1, 'bckgr', I) & ~INTERSECTIONUNDER(O1, 'pw', I) 
& ~INTERSECTIONUNDER(O1, 'gr', I) & ~INTERSECTIONUNDER(O1, 'r', I) 
>> ~ISTYPE(O1, 'ped', I) ^2

0.372: (O1 % O2) & DISTANCE(O1, O2, I) & ISTYPE(O2, 'ped', I) 
>> ISTYPE(O1, 'ped', I) ^2

0.278: (O1 % O2) & DISTANCE(O1, O2, I) & ISTYPE(O2, 'ped', I) & DISTANCE(O2, O3, I) 
& ISTYPE(O3, 'tr_sign', I) >> ISTYPE(O1, 'ped', I) ^2

0.002: (O1 % O2) & DISTANCE(O1, O2, I) & ISTYPE(O2, 'ped', I) 
& ISTYPE(O4, 'tr_signal', I) & DISTANCE(O2, O4, I) >> ISTYPE(O1, 'ped', I) ^2

0.899: ISTYPE(O3, 'tr_signal', I) & DISTANCE(O1, O3, I) & INTERSECTIONUNDER(O1, 'r', I) 
>> ISTYPE(O1, 'ped', I) ^2

0.246: ISTYPE(O3, 'tr_signal', I) & DISTANCE(O1, O3, I) & INTERSECTIONUNDER(O1, 'pw', I) 
>> ISTYPE(O1, 'ped', I) ^2

0.512: INTERSECTIONUNDER(O1, 'gr', I) & ISTYPE(O3, 'tr_signal', I) & DISTANCE(O1, O3, I) 
>> ISTYPE(O1, 'ped', I) ^2

0.213: ISTYPE(O3, 'tr_signal', I) & DISTANCE(O1, O3, I) 
& INTERSECTIONBEHIND(O1, 'bckgr', I) >> ISTYPE(O1, 'ped', I) ^2

1.4E-4: DISTANCE(O1, O3, I) & ISTYPE(O3, 'tr_sign', I) >> ISTYPE(O1, 'ped', I) ^2

0.317: ISTYPE(O4, 'tr_signal', I) & DISTANCE(O1, O4, I) >> ISTYPE(O1, 'ped', I) ^2

0.542: HASCOLOR(O3, 'white', I) & HASCOLOR(O3, 'blue', I) >> ISTYPE(O3, 'tr_sign', I) ^2

0.561: HASCOLOR(O3, 'yellow', I) >> ISTYPE(O3, 'tr_sign', I) ^2

0.841: HASCOLOR(O3, 'white', I) & HASCOLOR(O3, 'blue', I) >> ISTYPE(O3, 'tr_sign', I) ^2

0.181: HASCOLOR(O3, 'red', I) & HASCOLOR(O3, 'white', I) >> ISTYPE(O3, 'tr_sign', I) ^2

0.709: ~HASCOLOR(O3, 'white', I) & ~HASCOLOR(O3, 'blue', I) 
& ~HASCOLOR(O3, 'yellow', I) & ~HASCOLOR(O3, 'red', I) >> ~ISTYPE(O3, 'tr_sign', I) ^2

0.476: ~INTERSECTIONBEHIND(O3, 'bckgr', I) >> ~ISTYPE(O3, 'tr_sign', I) ^2

1.0 * ISTYPE(O, +Type, I) = 1.0 .
\end{verbatim}
\end{small}

\pagebreak
\subsection{Evaluation Plots}
\label{eval_plots}

\begin{figure}[h]
    \centering
    \begin{subfigure}[a]{0.8\textwidth}
        \centering
        \includegraphics[width=0.45\textwidth]{images/clean_run/ped_right.png}%
        \hfill
        \includegraphics[width=0.45\textwidth]{images/clean_run/ped_wrong.png}
        \caption{Scores distribution for the objects "pedestrians".}
    \end{subfigure}
    \vskip\baselineskip
    \begin{subfigure}[b]{0.8\textwidth}
        \centering
        \includegraphics[width=0.45\textwidth]{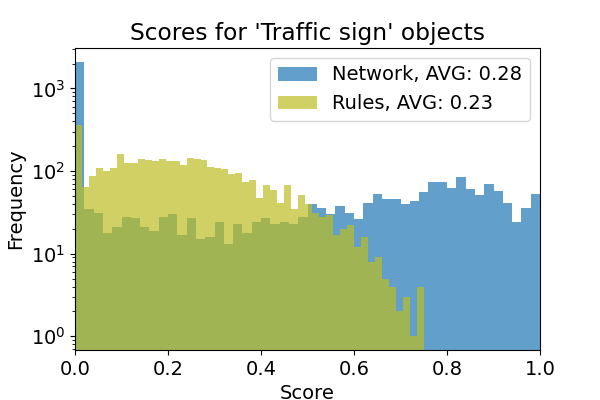}%
        \hfill
        \includegraphics[width=0.45\textwidth]{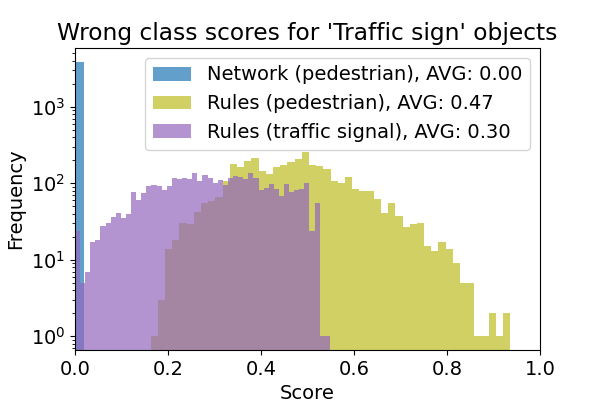}
        \caption{Scores distribution for the objects "traffic sign".}
    \end{subfigure}
    \vskip\baselineskip
    \begin{subfigure}[c]{0.8\textwidth}
        \centering
        \includegraphics[width=0.45\textwidth]{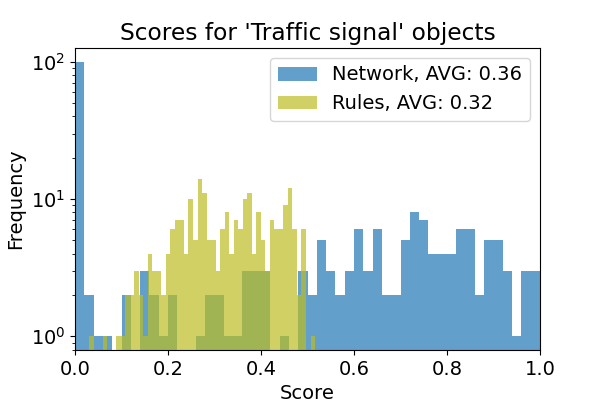}%
        \hfill
        \includegraphics[width=0.45\textwidth]{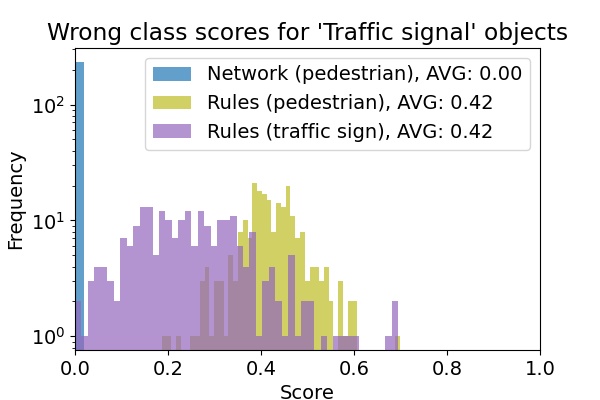}
        \caption{Scores distribution for the objects "traffic signal".}
    \end{subfigure}
    \caption{The scores of the network compared to the scores of the rules on the test dataset. It should be noted that the objects distribution is unbalanced in the test set: amount of objects "pedestrian" is the smallest, followed by "traffic signal". The largest number of objects are "traffic sign". The network can not distingusih between traffic signs or signals and we thus only report the (joined) result as ``tr\_sign'' for the wrong classes.}
    \label{fig:brightness5_preds}
\end{figure}

\pagebreak
\begin{figure}[h]
    \centering
    \begin{subfigure}[a]{0.8\textwidth}
        \centering
        \includegraphics[width=0.45\textwidth]{images/brightness5_run/ped_right.png}%
        \hfill
        \includegraphics[width=0.45\textwidth]{images/brightness5_run/ped_wrong.png}
        \caption{Scores distribution for the objects "pedestrians".}
    \end{subfigure}
    \vskip\baselineskip
    \begin{subfigure}[b]{0.8\textwidth}
        \centering
        \includegraphics[width=0.45\textwidth]{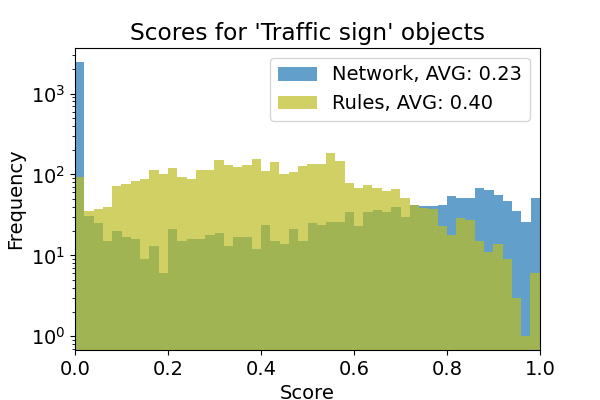}%
        \hfill
        \includegraphics[width=0.45\textwidth]{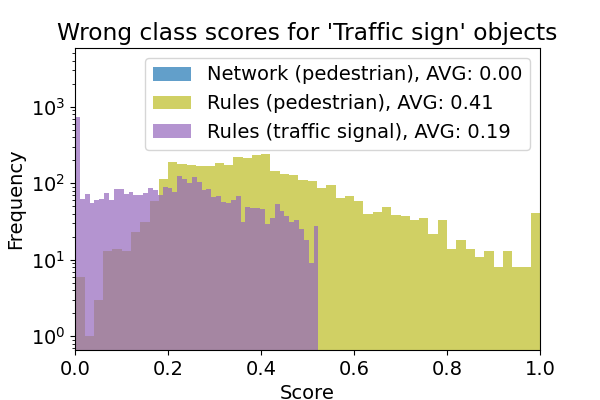}
        \caption{Scores distribution for the objects "traffic sign".}
    \end{subfigure}
    \vskip\baselineskip
    \begin{subfigure}[c]{0.8\textwidth}
        \centering
        \includegraphics[width=0.45\textwidth]{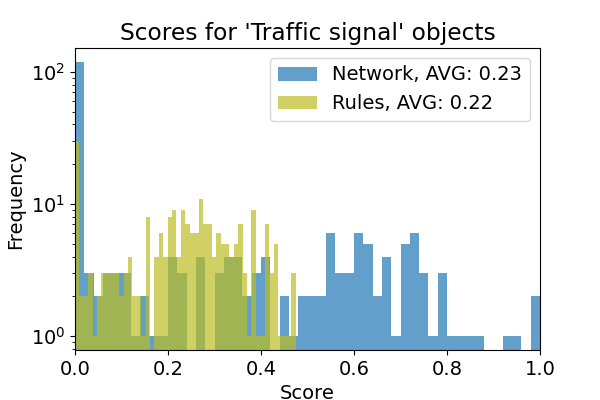}%
        \hfill
        \includegraphics[width=0.45\textwidth]{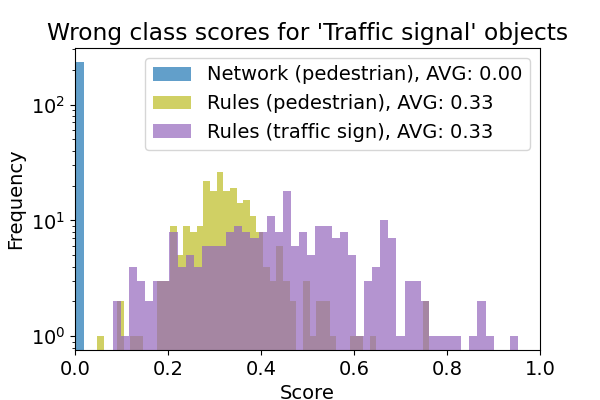}
        \caption{Scores distribution for the objects "traffic signal".}
    \end{subfigure}
    \caption{The scores of the network compared to the scores of the rules on the corrupted test dataset. The corruption selected here is "brightness" with the severity $5$. One can directly see that the network scores quality drops substantially while rules are nearly not affected by the corruption.}
    \label{fig:brightness5_preds2}
\end{figure}

\pagebreak
\subsection{Corruption Plots}
\label{all_corr_plots}

\begin{figure}[h]
    \centering
    \begin{subfigure}[a]{\textwidth}
        \centering
        \includegraphics[width=0.3\textwidth]{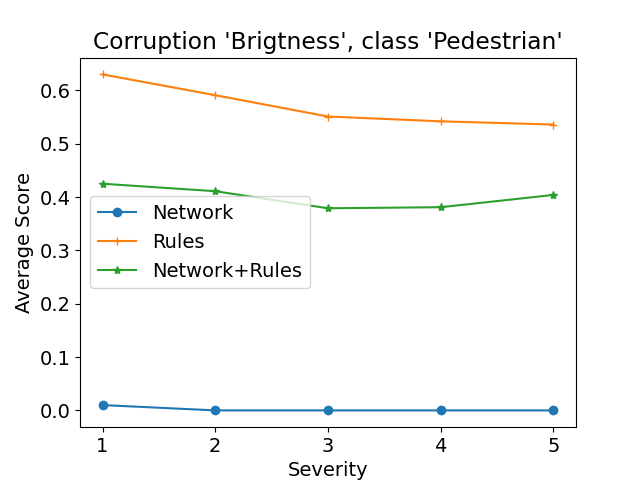}%
        \hfill
        \includegraphics[width=0.3\textwidth]{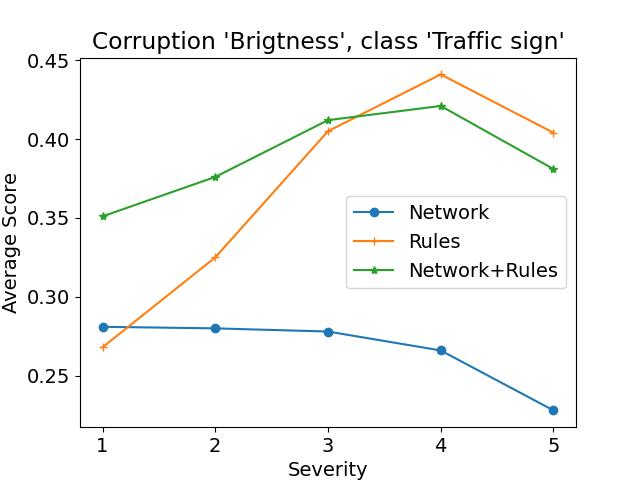}
        \hfill
        \includegraphics[width=0.3\textwidth]{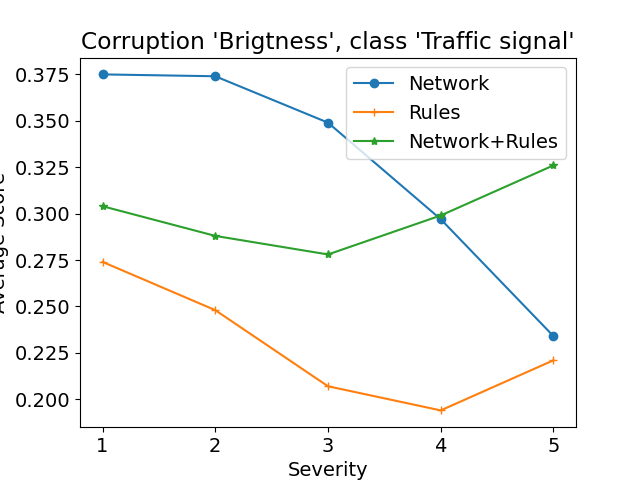}
        \caption{Comparison of the scores under increasing severity of ``brightness'' corruption.}
    \end{subfigure}
    \vskip\baselineskip
    \begin{subfigure}[b]{\textwidth}
        \centering
        \includegraphics[width=0.3\textwidth]{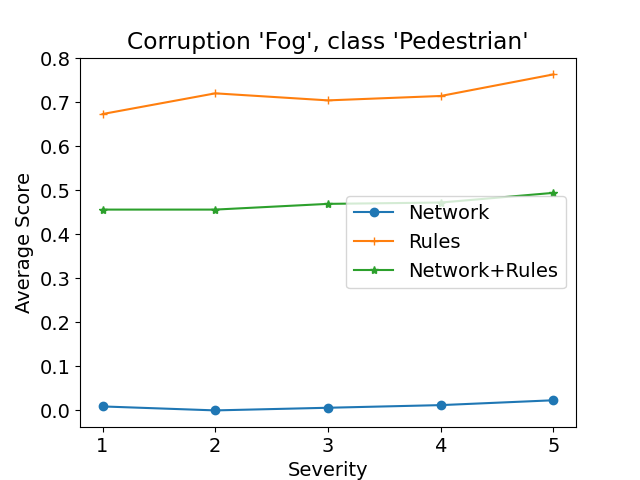}%
        \hfill
        \includegraphics[width=0.3\textwidth]{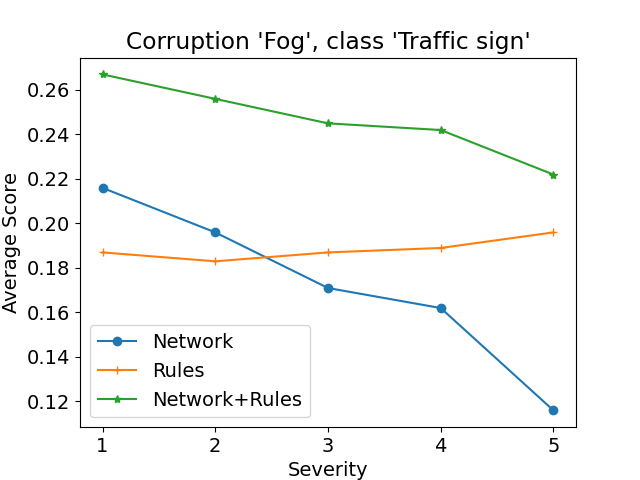}
        \hfill
        \includegraphics[width=0.3\textwidth]{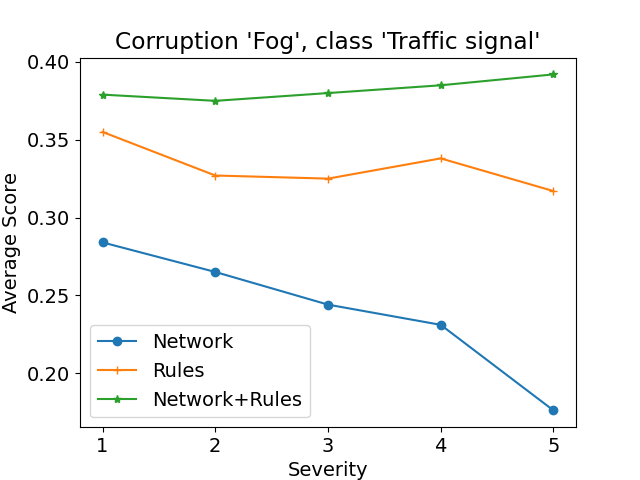}
        \caption{Comparison of the scores under increasing severity of ``fog'' corruption.}
    \end{subfigure}
    \vskip\baselineskip
    \begin{subfigure}[c]{\textwidth}
        \centering
        \includegraphics[width=0.3\textwidth]{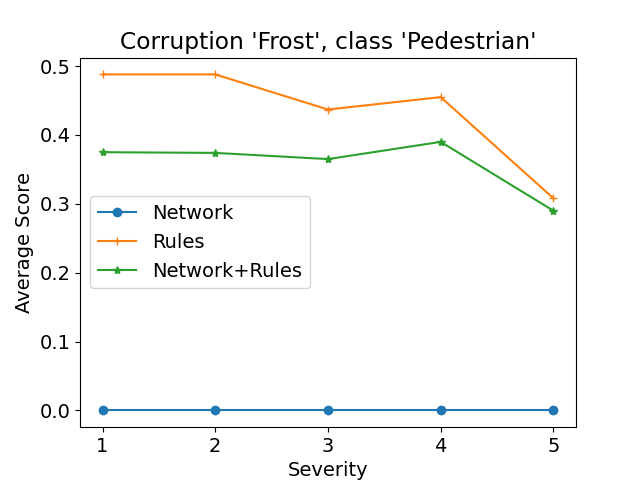}%
        \hfill
        \includegraphics[width=0.3\textwidth]{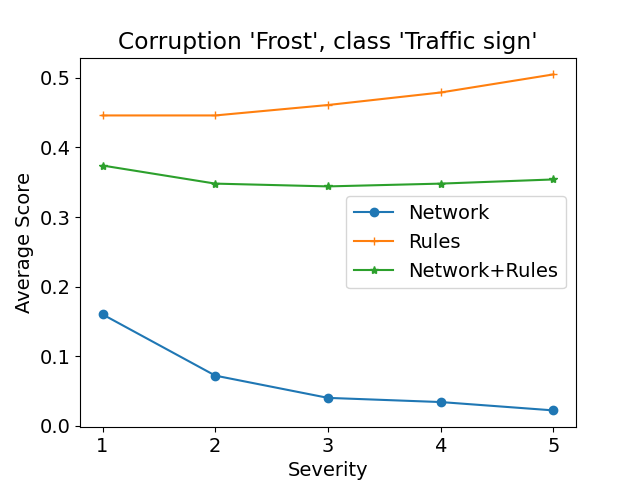}
        \hfill
        \includegraphics[width=0.3\textwidth]{images/severity_corruption/frost_comparison_tr_signal.png}
        \caption{Comparison of the scores under increasing severity of ``frost'' corruption.}
    \end{subfigure}
    \vskip\baselineskip
    \begin{subfigure}[d]{\textwidth}
        \centering
        \includegraphics[width=0.3\textwidth]{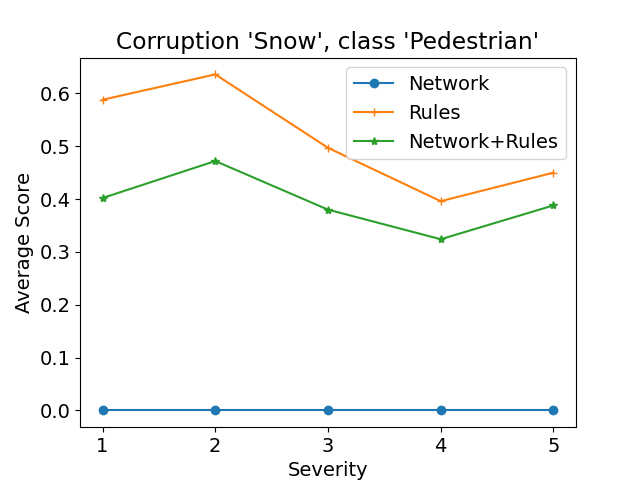}%
        \hfill
        \includegraphics[width=0.3\textwidth]{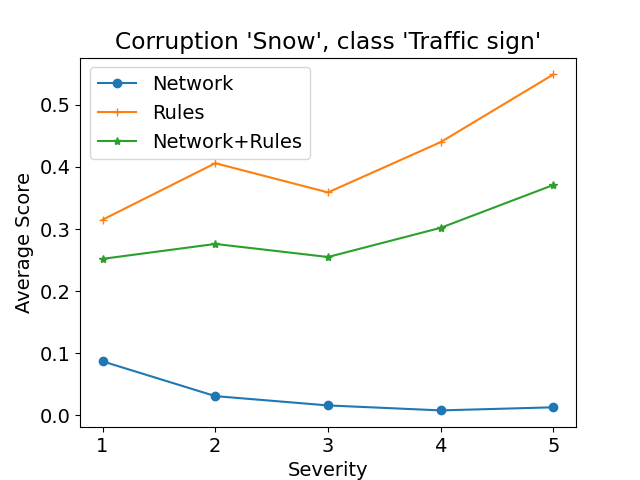}
        \hfill
        \includegraphics[width=0.3\textwidth]{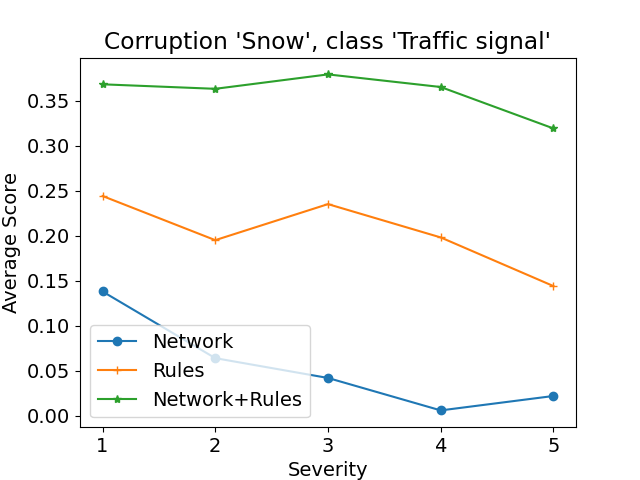}
        \caption{Comparison of the scores under increasing severity of ``snow'' corruption.}
    \end{subfigure}
    \caption{\small{Different type of corruptions with increasing severity. One can notice that corruptions can affect the rules scores: so brightness and snow add white color to the image, with that the rules for traffic signs are becoming more confident--based on the hasColor predicate. Another aspect is that amount of 'Pedestrian' objects is very little (~20 compared to ~2000 of 'Traffic sign') and because of this network mostly fails to distinguish them.}}
    \label{fig:all_severity_corruptions}
\end{figure}

\end{document}